\documentclass{article}

\usepackage{arxiv}

\usepackage[utf8]{inputenc} 
\usepackage[T1]{fontenc}    
\usepackage{hyperref}       
\usepackage{url}            
\usepackage{booktabs}       
\usepackage{amsfonts}       
\usepackage{nicefrac}       
\usepackage{microtype}      
\usepackage{lipsum}		
\usepackage{graphicx}
\usepackage{doi}
\usepackage{amsmath} 
\usepackage{caption}
\usepackage{tabularray}
\usepackage{adjustbox}
\usepackage[T1]{fontenc}
\usepackage[square,sort,comma,numbers]{natbib}

\title{CSANet: Channel Spatial Attention Network for Robust 3D Face Alignment and Reconstruction}

\date{}	

\author{ \href{https://orcid.org/0000-0000-0000-0000}{\includegraphics[scale=0.06]{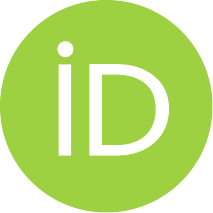}\hspace{1mm}Yilin Liu} \\
	Carnegie Mellon University\\
	\texttt{yilinli3@alumni.cmu.edu} \\
        \And
	\href{https://orcid.org/0000-0000-0000-0000}{\includegraphics[scale=0.06]{orcid.pdf}\hspace{1mm}Xuezhou Guo} \\
	Carnegie Mellon University\\
	\texttt{xuezhoug@alumni.cmu.edu} \\
	\And
	\href{https://orcid.org/0000-0000-0000-0000}{\includegraphics[scale=0.06]{orcid.pdf}\hspace{1mm}Xinqi Wang} \\
	Mount-Sheikh University\\
	\texttt{xinqiw@alumni.cmu.edu} \\
	\AND
	\href{https://orcid.org/0000-0000-0000-0000}{\includegraphics[scale=0.06]{orcid.pdf}\hspace{1mm}Fangzhou Du} \\
	Carnegie Mellon University\\
	\texttt{fdu@alumni.cmu.edu}
}

\renewcommand{\headeright}{}
\renewcommand{\undertitle}{}
\renewcommand{\shorttitle}{}

\hypersetup{
pdftitle={A template for the arxiv style},
pdfsubject={q-bio.NC, q-bio.QM},
pdfauthor={David S.~Hippocampus, Elias D.~Striatum},
pdfkeywords={First keyword, Second keyword, More},
}

\begin{document}
\maketitle

\begin{abstract}
	3D face reconstruction from monocular image has been a popular topic in the computer vision community. Recent breakthroughs are mainly facilitated by face alignment, which aims to locate face vertices on a registered model. Although existing research works successfully tackle accuracy problems caused by large head pose, they still fall short when the faces are partially occluded, or in large light changing condition. Motivated by these challenges, this project proposes an end-to-end 3D face alignment and reconstruction network. The backbone of
our model is built by BottleNeck structure via Depthwise Separable Convolution. In order to improve the model’s performance in extreme conditions, we integrate Coordinate Attention mechanism and Spatial Group-wise Enhancement to extract more representative features in face regions. For more stable training process and better convergence, we jointly use Wing loss and the Weighted Parameter Distance Cost (WPDC) to learn parameters for 3D Morphable model (3DMM) and 3D
vertices. From the results, our proposed model outperforms all baseline models both quantitatively and qualitatively. Specifically, the face alignment accuracy is improved by 7.44\% on AFLW and 10.16

\end{abstract}


\section{Introduction}
During the past few years, 3D image processing is gaining more attention due to the limitations of
2D image application. Many research fields benefit from the booming of 3D images, such as robotics,
security and bio-medical. 3D images can associate with 2D images, while providing more enriched
information (e.g. space directions). Among all 3D image related tasks, 3D face reconstruction is
considered as one of the most challenging ones, since using scanning device to obtain ground-truth
3D faces can be expensive and time-consuming. Therefore, reconstructing 3D face models from a 2D
face image has been widely explored, as an alternative to synthesize accurate 3D faces with limited
budget.
Generally, 3D face reconstruction can be decomposed into two steps, the initial face alignment step
and the face detail synthesis step \cite{Li2021}. The initial modeling mainly involves face alignment, which
focuses on locating facial vertices of certain face model, and face synthesis aims at recovering 3D
facial geometry and details from the input image. In particular, the accuracy of the alignment module
has a significant effect on the overall performance of the reconstruction model.
Our project aims to improve the performance of 3D face alignment frameworks. Furthermore, based
on the alignment results, we can generate more precise 3D face reconstruction results. Initially, we
re-implemented the 3D face reconstruction model 3DDFA \cite{8122025}. The authors proposed to use the Pose
Adaptive Feature (PAF) and Projected Normalized Coordinate Code (PNCC) to encode human face
features. However, these two processes turned out to involve tremendous computation, which is not
efficient during evaluation and interference, thus making it impossible to be applied in real-world
applications. Besides, when the face image is partially occluded (e.g. by hair or sunglasses) or in
large lighting change conditions, 3DDFA and some other proposed methods struggle to generate
desirable alignment and reconstruction results.
To address these problems, we improve the baseline model in three aspects. Firstly, inspired by the
MobileNet \cite{DBLP:journals/corr/HowardZCKWWAA17} structure and DAMDNet \cite{DBLP:journals/corr/abs-1908-11821}, we adopt Depth-wise Separable Convolution instead
of normal convolution for efficient computation. Then, we integrate the state-of-the-art attention
mechanisms along both the spatial and channel dimensions to deal with faces with occlusions and
lighting condition changes. Lastly, two different losses are jointly used to learn parameters for the
projected 3D Morphable Model (3DMM) \cite{10.1145/3596711.3596730, 1227983}. Extensive experiments are conducted on the AFLW
and AFLW2000-3D benchmark and the performance of the proposed model is then compared with
that of other baseline models.

\section{Related Work}
\label{sec:headings}

\subsection{Face Alignment}
Face alignment aims to accurately locate facial vertices. It can be considered as a model parameter
optimization problem \cite{DBLP:journals/corr/abs-2106-03021}. In the past, face alignment focuses on detecting sparse 2D facial landmarks
in the image plane. Many traditional works \cite{sauer2011accurate, SUNG2009359} are based on the Active Appearance Models
(AMM) \cite{cootes2001active} or Active Shape Models (ASM) \cite{cootes1995active}, which only represent shape/texture variations with
linear subspace. To better model non-linear pose variations, such as face rotation, some improvements
are proposed to add non-linearity into face model \cite{romdhani1999multi, zhou2005bayesian}, or into the model fitting formulation \cite{saragih2007nonlinear}.
Despite the introduction of non-linearity to improve the representative ability of 2D face alignment
methods, there are still limitations on detecting invisible landmarks (e.g. occlusions, profile view),
since 2D points lose geometric correspondence with actual locations. Therefore, 3D face alignment
approaches have become the mainstream research in recent years. 3D face alignment can be separated
into 2 strategies. (1) Bypass constructing and fitting face models. \cite{DBLP:journals/corr/abs-1803-07835, DBLP:journals/corr/JacksonBAT17} directly regress all 3D
point coordinates by a fully convolutional neural network. However, the resolution of resulted faces
depends on the feature map size, and it is not time-effective during inference \cite{DBLP:journals/corr/abs-2009-09960}. (2) Implement
parameter regression of a certain 3D face model \cite{8571265,liu2017dense,mcdonagh2016joint,DBLP:journals/corr/TranHMM16,DBLP:journals/corr/abs-1804-01005} to obtain a full 3D structure, in order
to guide the actual landmark localization. Among all 3D face representing methods, 3D Morphable
Model (3DMM) \cite{blanz2023morphable, blanz2003face} is the most widely-used one. 3DMM is a vector space of linear combinations
of objects’ shapes (i.e. vertex coordinates) and textures (i.e. RGB colors). It applies Principle
Component Analysis (PCA) to 3D face scans. The non-linear face transformations are separate from
linear subspaces, which models linear transformations for shapes or expressions.

\subsection{Face Reconstruction}

There are two main categories of 3D face reconstruction: model-based and model-free approaches.
The model-based reconstruction methods are highly related with 3D dense face alignment if the face
model is well registered. Initially, earlier studies \cite{huber2015fitting, lee2012single, paysan20093d, 1467550} reconstruct 3D faces by regressing
3DMM parameters and as solving a nonlinear optimization problem. Later, works \cite{jourabloo2016large, tran2017regressing, 8954451} based
on Convolution Neural Networks (CNNs) started to replace traditional parameter-learning methods
with more accurate results. Since the reconstruction geometry is limited by linear bases of 3DMM,
some works proposed to use non-linear models. For instance, \cite{tran2019learning} learns a nonlinear 3DMM decoder
to reconstruct 3D faces from in-the-wild face images, instead of face scans. Similarly, \cite{DBLP:journals/corr/abs-1808-05323} learn
a nonlinear 3DMM from diverse sources, including RGB-D images. Some other works \cite{fan2020dual, DBLP:journals/corr/abs-1803-07835,DBLP:journals/corr/abs-1807-10267}
directly estimate 3D face geometry without depending on parametric models. These methods do not
require accurate alignment or dense correspondence. \cite{DBLP:journals/corr/JacksonBAT17} adopts a volumetric representation of 3D face shape and directly regressed the voxels.

\subsection{Attention}
Attention mechanism was first introduced in natural language processing field. Due to its effectiveness
in feature extraction, it has also been deployed in various visual-based research works \cite{Hu_2018_CVPR, DBLP:journals/corr/abs-1807-06514, woo2018cbam}.
The visual-based attention mechanism has shown the potential ability in refining extracted features
obtained by convolutions. Specifically, existing attention modules mainly refines features along
the channel and spatial dimensions. The channel attention determines which features are more
important, while spatial attention focuses on which portion of the image can be more informative.
By applying attention mechanism, the model can adaptively ignore the less important information
by enhanced feature representation. In this project, we improve the baseline model by adding two
attention mechanisms to capture the relation between channel and spatial feature groups. One is
Spatial Groupwise Enhancement(SGE) \cite{li2019spatial} added inside each layer to enhance spatial group-wise
feature capture, another is Coordinate Attention (CA) \cite{hou2021coordinate} added between each layer to aggregate features along two spatial coordinate directions respectively.

\section{Baseline Implementation}
Among existing popular frameworks \cite{DBLP:journals/corr/abs-1803-07835, DBLP:journals/corr/JacksonBAT17, DBLP:journals/corr/abs-1807-10267, DBLP:journals/corr/abs-1808-05323,8122025}, we chose 3DDFA \cite{8122025} as the face alignment
baseline model to deal with large pose 3D regression. While being one of the state-of-the-art solutions
for large-pose face alignment, 3DDFA is also one of the fundamental models, which is frequently
cited by other 3DDFA-based variations or novel face alignment and reconstruction techniques for
comparison and benchmark performance evaluation.

\subsection{3D Morphable Model}

3DMM \cite{10.1145/3596711.3596730, 1227983} is a way of representing face shape with the formula:
\begin{equation} \label{eq:1}
S = \overline{S} + A_{id}\alpha_{id} + A_{exp}\alpha_{exp}
\end{equation}
, where $\overline{S}$ is the average face, $A_{id}$ is the principle component of training 3D face scans, $\alpha_{id}$ is the shape parameter, $A_{exp}$ is the principle axes trained on the offsets between expression scans and neutral scans, and $\alpha_{exp}$ is the expression parameter. After attaining the 3D face model, it can be projected onto 2D plane with the transformation formula:
\begin{equation} \label{eq:2}
V(p) =  f * Pr * R * (\overline{S} + A_{id}\alpha_{id} + A_{exp}\alpha_{exp}) + t_{2d}
\end{equation}
, where $Pr = \begin{pmatrix}
1 & 0 & 0 \\
0 & 1 & 0 
\end{pmatrix}  $
represents the orthographic projection matrix, $f$ if the scale factor, $R$ is the rotation matrix and $t_{2d}$ is the translation vector.

\subsection{Network Structure and Cost Function}
\begin{figure}
\captionsetup{justification=centering}
    \centering
    \includegraphics[width=0.6\linewidth]{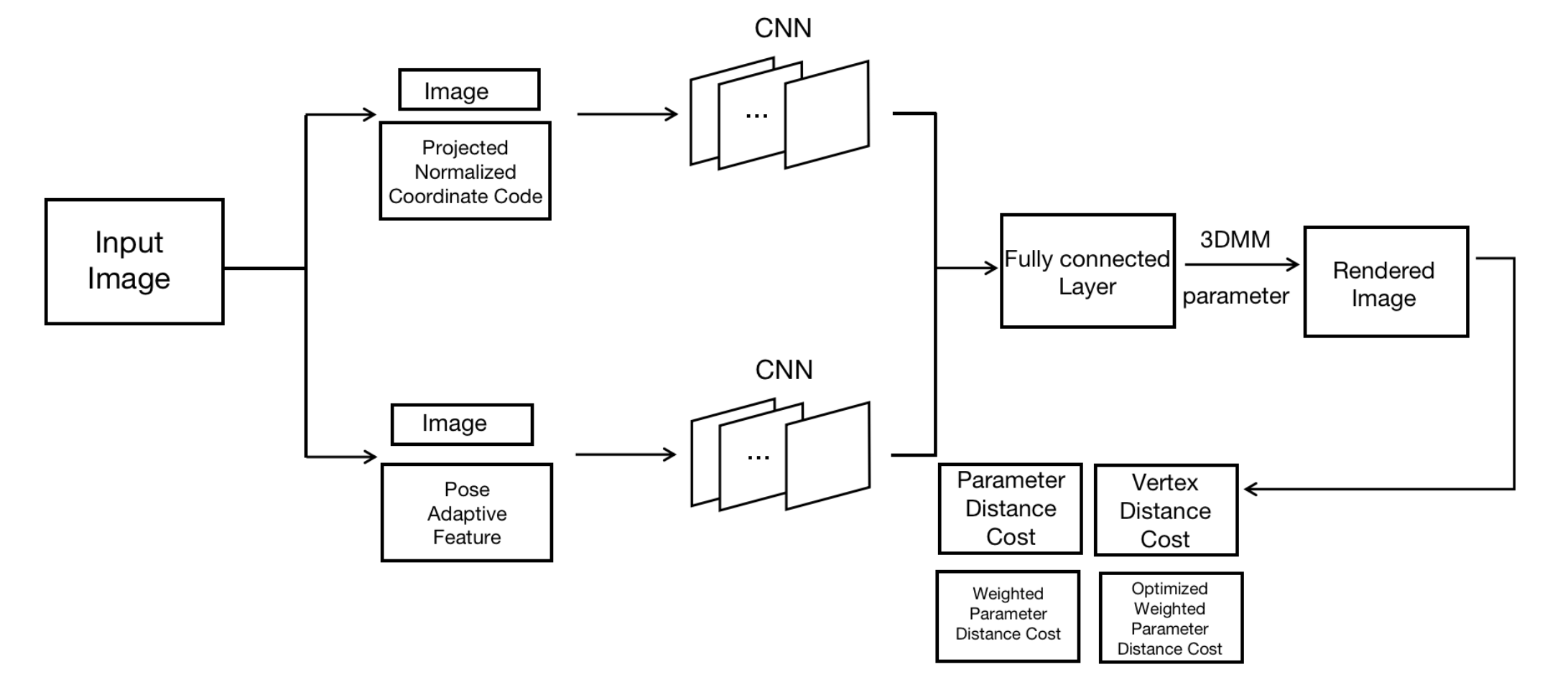}
    \caption{3DDFA Pipeline}
    \label{fig:fig1}
\end{figure}

In 3DDFA, the main pipeline is a two-stream CNN network designed with unified network structure
across the cascade \cite{8122025}, as shown in Figure \ref{fig:fig1}. For each iteration, the network first constructs the
Projected Normalized Coordinate Code (PNCC) and the Pose Adaptive Feature (PAF) \cite{8122025} based on current parameters for input images. The calculated PNCC and PAF data, combined with the original
face images, are used to train the two streams. To be more specific, the first CNN stream is trained with input images combined with PNCC values, while the second CNN stream is trained with input
images combined with PAF values. The outputs of the two CNN streams are concatenated and passed through fully connected layers to generate the predicted parameter update values $\Delta p$ that will be used
to update the network \cite{8122025}.

In the original work, three cost functions are discussed: Parameter Distance Cost (PDC), Vertex Distance Cost (VDC), and Weighted Parameter Distance Cost (WPDC) \cite{8122025}. Among the three cost functions, the Weighted Parameter Distance Cost (WPDC) proposed in the paper showed the
best performance as the loss function. WPDC aims at shrinking the difference between the current parameters $p_{0} + \Delta p$ and the ground-truth $p_{g}$ while assigning different weights to different parameters to emphasize their importance \cite{8122025}.

\subsection{Implementation Results}
Our re-implementation of 3DDFA baseline model is based on the source code published by the
original authors \cite{DBLP:journals/corr/abs-2009-09960}. The reason that we re-implemented the original proposed method with source
code published by original authors is that the published implementation does not feature the same
techniques stated in the original paper. The authors also continuously updated their implementation
with some strategies not discussed in the paper, and optimized their design targeting inference
efficiency instead of accuracy by reducing the two-stream CNN structure to single-stream with more
advanced CNN structure. Since we could not verify the impact of these changes on face alignment
and reconstruction accuracy, we discarded efficiency-targeting changes not described in the original paper and re-implement the two-stream CNN as the paper proposed. This ensure accurate comparison
among methods.

Test results are obtained by training from scratch with the same data provided by the original authors,
and are compared with that obtained by model trained with released source code, released pre-trained
model, as well as that the paper reported values as shown in Table \ref{tab:table1}. As the results show, our
re-implemented version of the two-branch structure performed worse on the benchmark compared
to the published pre-trained models and the results reported in the paper, but slightly better than the
model we obtained strictly following the source code and training procedures release by the authors.
One possible reason for this is variance and lack of parameters tuning. We got better performance
with PDC and worse with WPDC comparing to the reported values. Still, as the original authors
stated in the documentation of their publish source code, accuracy may vary depending on selection
of training and testing datasets \cite{DBLP:journals/corr/abs-2009-09960}, and our test results are within the error range. Therefore, we
consider our implementation as valid.

\begin{table}
\captionsetup{justification=centering}
	\caption{Comparison of the Normalized Mean Error (NME\%) of face alignment results \\ on AFLW2000-3D using WPDC loss}
	\centering
	\begin{tabular}{llll}
		\toprule
		\cmidrule(r){1-2}
		Our re-Implementation & Published Source Code & Published Pre-trained & Paper \\
		\midrule
		$4.759$ & $4.823$ & $4.252$ $\pm$ $0.996$ & $4.05$\\
		\bottomrule
	\end{tabular}
        \label{tab:table1}
\end{table}

\section{Channel Spatial Attention Network}
\subsection{Model Structure}
\begin{figure}
\captionsetup{justification=centering}
    \centering
    \includegraphics[width=0.6\linewidth]{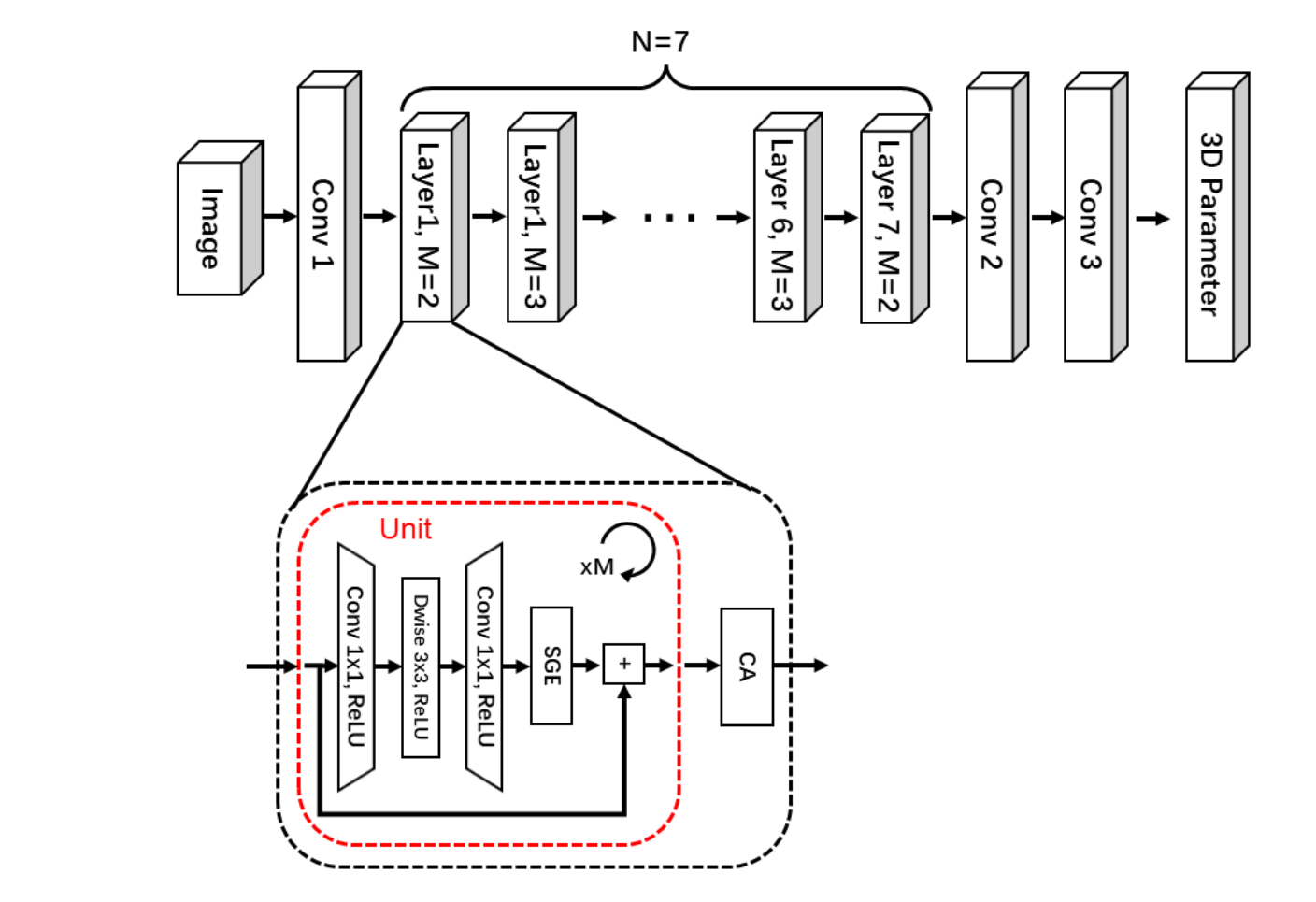}
    \caption{The proposed model structure}
    \label{fig:fig2}
\end{figure}

The network architecture for our model is shown in Figure \ref{fig:fig2}. The main pipeline is inspired by \cite{DBLP:journals/corr/abs-1908-11821}.
The input to our model is a single RGB face image, and the output is a 62-dimensional 3D parameter vector, including 40-dimensional identity parameter vector, 10-dimensional expression parameter vector and 12-dimensional pose vector. There are 7 main layers in the model backbone. Each layer is a cascade of several units. Each unit is shown in the red box in Figure \ref{fig:fig2}. It is a bottleneck structure followed by a Spatial Group-wise Enhancement (SGE) module. Between each layer, we add another attention block: coordinate attention. Both attention mechanisms and the cost function will be discussed in more details in the following sections.

\subsection{Attention Modules}
\subsubsection{Spatial Group-wise Enhance Module}
\begin{figure}
\captionsetup{justification=centering}
    \centering
    \includegraphics[width=0.6\linewidth]{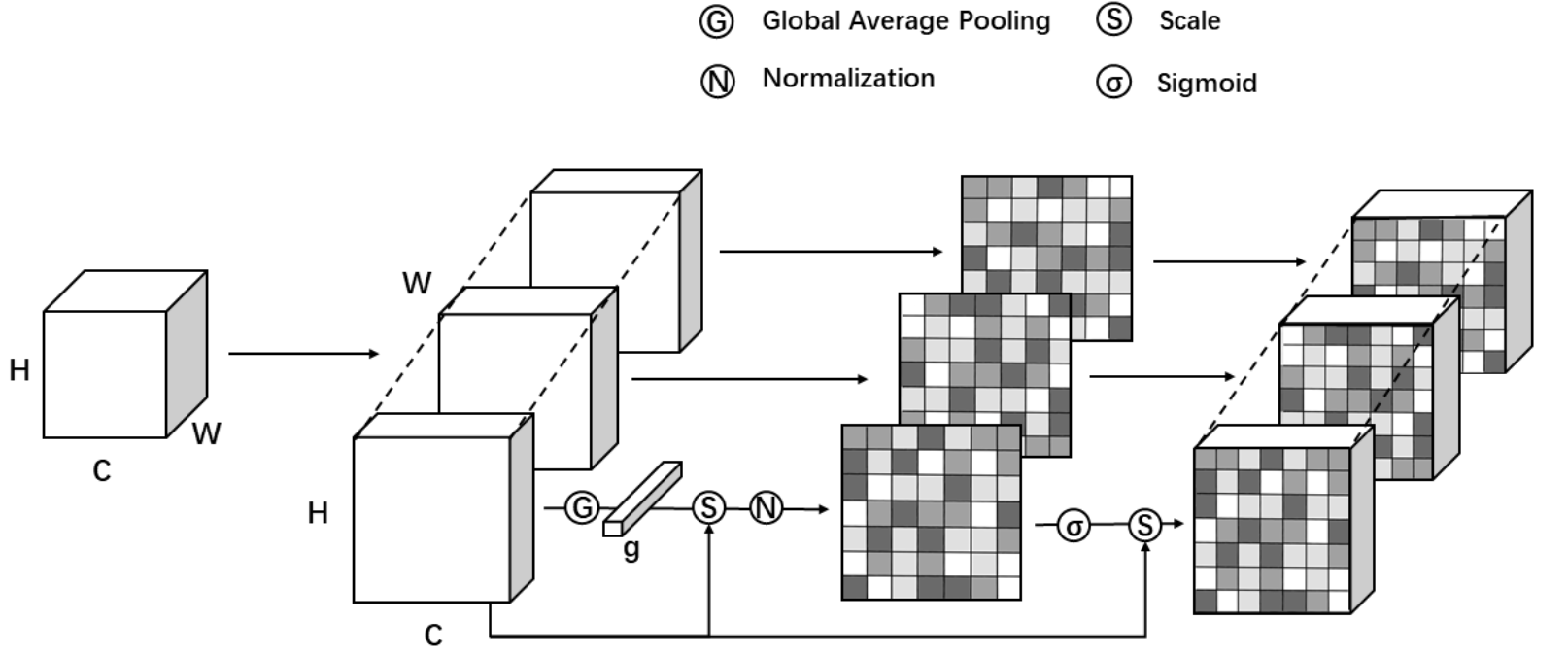}
    \caption{Spatial Group-wise Enhance Module(SGE)}
    \label{fig:fig3}
\end{figure}

In face analysis tasks, CNNs are widely used to extract semantic features that can be grouped along the channel-dimension as sub-features \cite{li2019spatial}. However, these approaches overlooked the spatial dimension of the convolution feature maps. In fact, it would be desired to generate the corresponding features from the correct location in the original image for a particular semantic group. Still, this can be made challenging by noises in the background and image patterns. To account for the spatial dimension, we employed the Spatial Group-wise Enhance (SGE) \cite{li2019spatial} module shown in Figure \ref{fig:fig3} to strengthen the the spatial distribution of facial semantic features. This attention mechanism treats each feature layer as a sub-feature group, and generates an attention factor based on the similarity of global and local features within each group. First, the feature layers will be grouped along the channel dimension and each group is examined. Ideally, this group contains features representing a specific semantic response such as eyes or mouths, but this is obscured by the noises in the actual image. To account for this instability, it is better to use the entire group space to improve the learning of the critical regions. The global statistical feature can be obtained as
\begin{equation} \label{eq:3}
g = F_{gp}(X) = \frac{1}{m} \sum^{m}_{i=1}x_{i} 
\end{equation}

where $F_{gp}(\cdot)$ is a spatial averaging function. With the global feature, we calculate the initial attention mask by spatially comparing each local feature group $x_{i}$ with the global pooling feature $g$, which
can be obtained by the dot product that measure the similarity between the them. Next, the final attention mask $a_{i}$ is obtained by a sigmoid operation on the normalized initial attention mask, and the enhanced features $\hat{x}_{i}$ can be generated as
\begin{equation} \label{eq:4}
\hat{x}_{i} = x_{i} \cdot \sigma(a_{i})
\end{equation}

\subsubsection{Coordinate Attention Module}
\begin{figure}
    \centering
    \includegraphics[width=0.7\linewidth]{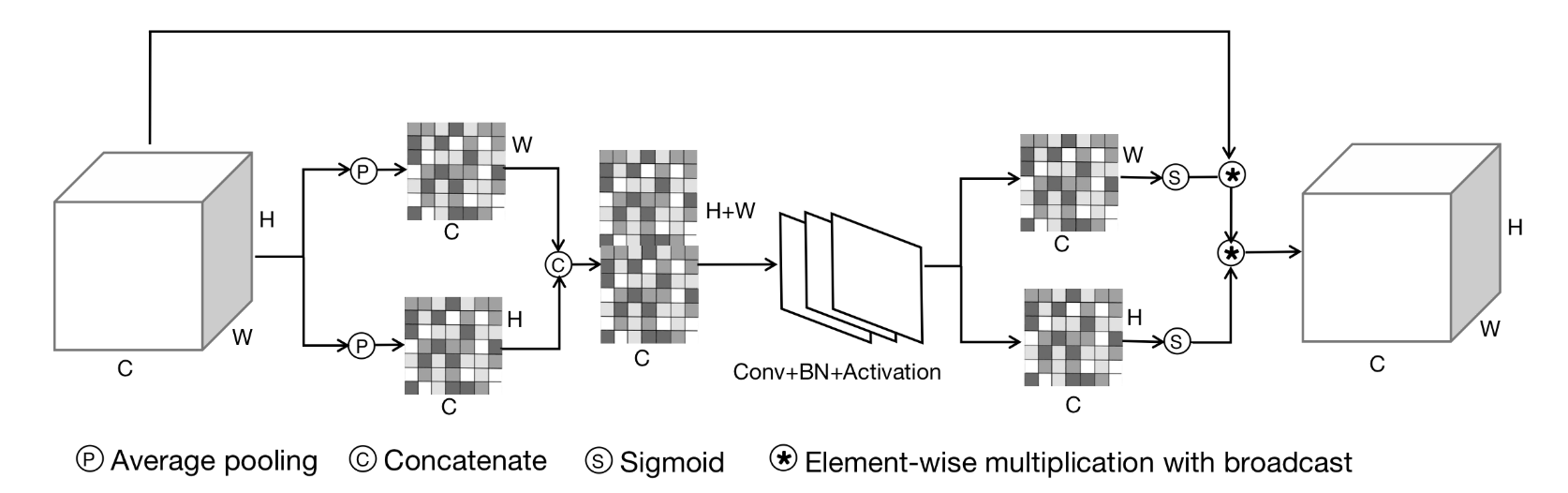}
    \caption{Coordinate Attention Module(CA)}
    \label{fig:fig4}
\end{figure}

To better represent precise positional information while also effectively capturing inter-channel relationships, Coordinate Attention \cite{hou2021coordinate} blocks illustrated in Figure \ref{fig:fig4} are introduced into the network. Specifically, it will squeeze spatial content by two pooling kernels along the horizontal and vertical coordinate to encode each channel information. This will yield a pair of feature maps, which aggregate features along two spatial directions respectively. After encoding the positional information, we generate intermediate feature maps by applying $1\times1$ convolution $F_{1}$ on the concatenated feature maps
as
\begin{equation} \label{eq:5}
f = \delta(F_{1}([z^{h}, z^{w}]))
\end{equation}
where $z^{h}$ and $z^{w}$ are the generated feature maps and $\delta$ is a non-linear Swish activation function, which is simply $x \ast sigmoid(x)$. $f$ is then split along the spatial dimension into $f^{h}$ and $f^{w}$ and
processed again with $1\times1$ convolutions followed by sigmoid activation, giving
\begin{equation} \label{eq:6}
g^{h} = \sigma(F_{h}(f^{h}))
\end{equation}
\begin{equation} \label{eq:7}
g^{w} = \sigma(F_{w}(f^{w}))
\end{equation}

Lastly, the attentions are applied to the input of the block as
\begin{equation} \label{eq:8}
y_{c}(i,j) = x_{c}(i,j) \times g^{c}_{c}(i) \times g^{w}_{c}(j)
\end{equation}

\subsection{Loss Function}
In this project, we adopt a joint loss \cite{DBLP:journals/corr/abs-1908-11821}, which is composed of the Wing Loss \cite{DBLP:journals/corr/abs-1711-06753} and the WPDC loss \cite{8122025}. The WPDC loss, as discussed before, reduces the differences between the generated face model parameters and the target parameters; the Wing loss improves the accuracy of CNN-based facial landmark localisation. It deals with both large errors that occur at the beginning of the training process and smaller errors during later phases. They are combined as
\begin{equation} \label{eq:9}
loss_{merged} = loss_{wing} + 0.5loss_{wpdc}
\end{equation}

\section{Experiments}
\subsection{Implementation}
Our model is implemented using the PyTorch library. The training and validation processes are implemented on Amazon EC2 g4dn.2xlarge instances. The model was trained for 50 epochs, approximately 25 hours to achieve the obtained results. For 3DDFA baseline re-implementation, on the same device, it costs about 50 hours to train. Therefore, our proposed model is two times faster in terms of training speed. We used Stochastic Gradient Descent (SGD) as the parameter optimizer with and initial learning rate at 0.02, and reduced it by a factor of $0.6$ after no significant improvements
for 4 epochs. Details of training/validation datasets and benchmark evaluation are discussed in the following sections.

\subsection{Dataset and Data Transformation}
We used 300W-LP as the training dataset. It is a synthesized face alignment dataset containing 122,450 samples with large head poses, which are generated by profiling the face images in the original 300W dataset. 300W dataset consists of multiple re-annotated in-the-wild datasets with 68 facial landmarks. 

We evaluated the trained model on the AFLW and the AFLW2000-3D datasets. AFLW is a large-scale collection of annotated face images (21,080 in-the-wild faces) gathered from the web, exhibiting a large variety in appearance (e.g., pose, expression, gender). Each image is annotated with 21 landmarks. AFLW2000-3D is developed from AFLW dataset. It contains the ground truth 3D face scans and the corresponding 68 landmarks for the first 2,000 AFLW samples. This dataset fills in the gap between mapping from 2D landmarks to its 3D scans. 

In addition, images from all datasets are pre-processed before feeding to the network. Each image is cropped according to the region of interest, which is the bounded face region in the original image. The standard size of cropped images are $120\times120$. During the benchmark evaluation, images are also categorized by yaw angles, representing small, medium and large poses respectively. For example, yaw angle within $[0^{\circ}, 30^{\circ}]$ can be regarded as small head movement.

\subsection{Evaluation Metric}
Normalized Mean Error (NME) is used to evaluate the face alignment performance of the network. It is defined as the normalized mean Euclidean distance between corresponding points in the predicted result $p$ and the ground truth $\hat{p}$.
\begin{equation} \label{eq:10}
NME = \frac{1}{N}\sum_{i=1}^{N}\frac{\Vert p_{i} - \hat{p}_{i}\Vert_{2}}{d}
\end{equation}
where the normalization factor $d$ is the bounding box size of ground truth landmarks.

For face reconstruction results, we fitted the obtained 3D parameters into 3DMM and visualize it in 3D vector space. The face reconstruction results are consistent with face alignment landmarks and further discussed in the Results section.

\section{Results}
The model performance is evaluated both quantitatively and qualitatively. Numerical results are
shown in Table \ref{tab:table2}. We compare a series of 3DDFA models with different cost functions, one of the
state-of-the-art face reconstruction model named DAMDNet and our proposed model. The baseline
3DDFA model is the published version, which is trained by Weighted Parameter Distance Cost
(WPDC) loss and fine-tuned on Vertex Distance Cost (VDC), denoted as 3DDFA-WPDC-VDC.
Other 3DDFA models are trained by Parameter Distance Cost (PDC), VDC and WPDC separately,
denoted by the name of adopted loss function in the table. Besides, we also include the evaluation
of the proposed model without SGE blocks as a comparison, to validate the effectiveness of the
incorporation of SGE into the model.From the result, our proposed network outperforms other
baseline models, especially in large pose conditions. Compared with the chosen baseline model
3DDFA, the face alignment accuracy is improved by 7.44\% on the AFLW dataset and 10.16\% on the AFLW2000-3D dataset.

\begin{table}
\captionsetup{justification=centering}
\centering
\caption{The NME(\%) of face alignment results on AFLW and AFLW2000-3D with the first and the second best results highlighted.}
\begin{adjustbox}{width=6in,center}
\begin{tblr}{
  cell{1}{2} = {c=5}{},
  cell{1}{7} = {c=5}{},
  vline{2-3} = {1}{},
  vline{2,7} = {2-10}{},
  hline{2-3} = {-}{},
}
         & AFLW Dataset &                &                &                &       & AFLW2000-3D Dataset &                &                &                &       \\
Method         & ${[}0^{\circ},30^{\circ}]$      & ${[}30^{\circ},60^{\circ}]$    & ${[}60^{\circ},90^{\circ}]$    & Mean           & Std   & ${[}0^{\circ},30^{\circ}]$          & ${[}30^{\circ},60^{\circ}]$    & ${[}60^{\circ} ,90^{\circ} ]$  & Mean           & Std   \\
3DDFA-PDC      & $6.937$           & $8.441$          & $9.175$          & $8.184$          & $0.932$ & $5.254$               & $6.673$          & $8.123$          & $6.683$          & $1.172$ \\
3DDFA-VDC   & $5.756$           & $7.222$          & $8.072$          & 7.017          & $0.957$ & $4.040$               & $5.207$          & $6.654$          & $5.300$          & $1.069$ \\
3DDFA-WPDC     & $5.282$           & $6.606$          & $7.464$          & $6.451$          & $0.898$ & $3.658$               & $4.857$          & $5.962$          & $4.826$        & $0.941$ \\
3DDFA-WPDC-VDC & $\textbf{4.488}$  & $5.381$          & $6.334$          & 5.401          & $0.754$ & $3.099$               & $4.172$          & $5.486$          & $4.252$          & $0.976$ \\
3DDFA-ours     & $5.206$           & $6.496$          & $6.975$          & $6.226$          & $0.747$ & $3.897$               & $4.783$          & $5.788$          & $4.823$          & $0.733$ \\
DAMDNet        & $4.539$           & $\textbf{5.209}$ & $\textbf{6.028}$ & $\textbf{5.199}$ & $0.682$ & $\textbf{2.907}$      & $\textbf{3.830}$ & $\textbf{4.953}$ & $\textbf{3.897}$ & $0.837$ \\
Ours-CA        & $4.978$           & $6.072$          & $6.885$          & 5.978          & $0.781$ & $3.440$               & $4.300$          & $5.376$          & $4.372$          & $0.792$ \\
Ours           & $\textbf{4.254}$  & $\textbf{4.947}$ & $\textbf{5.796}$ & $\textbf{4.999}$ & $0.630$ & $\textbf{2.914}$      & $\textbf{3.738}$ & $\textbf{4.809}$ & $\textbf{3.820}$ & $0.776$ 
\end{tblr}
\end{adjustbox}
\label{tab:table2}
\end{table}

\begin{figure}
    \centering
    \includegraphics[width=0.6\linewidth]{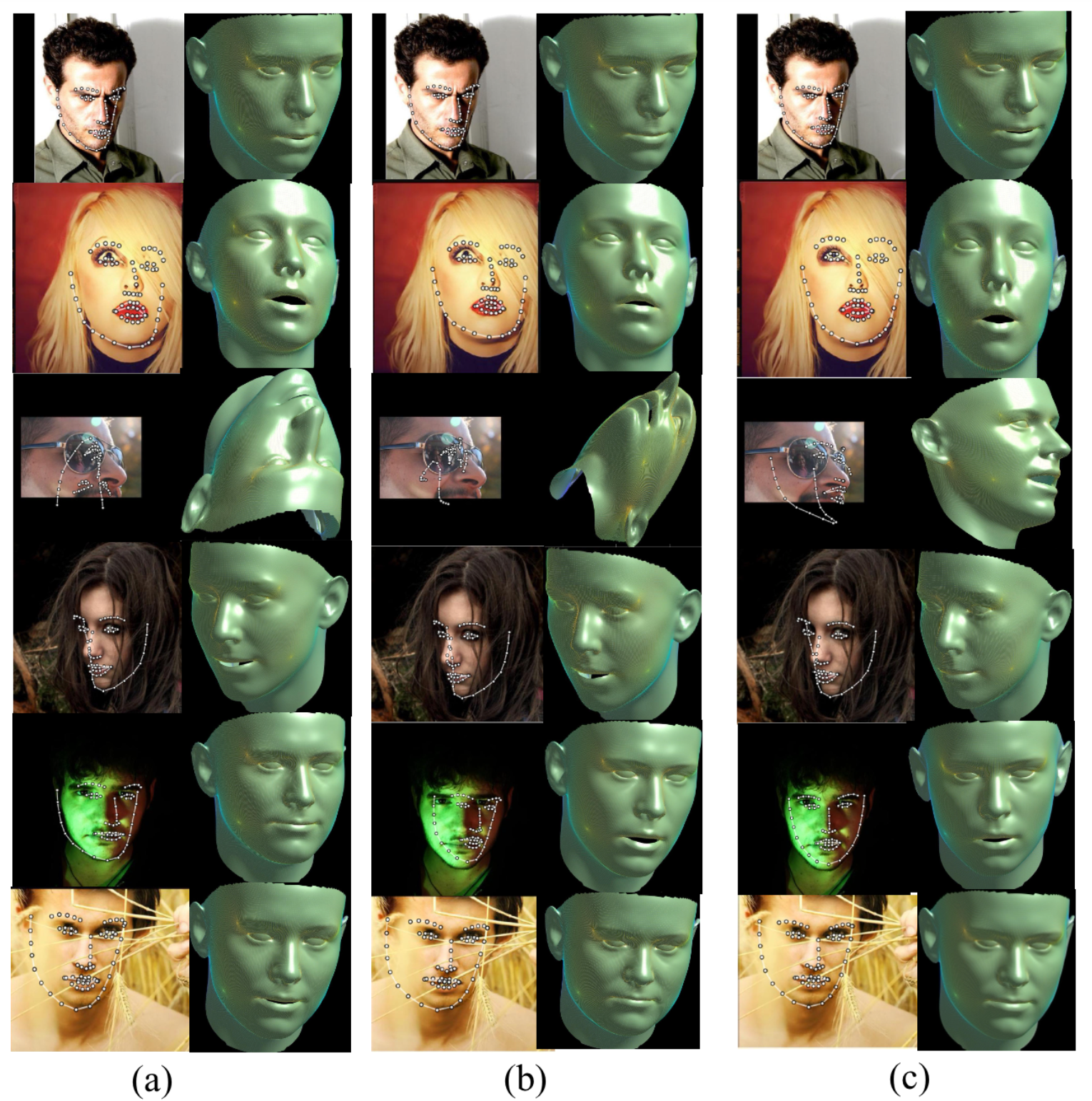}
    \caption{Qualitative Visual Results of 3DDFA(a), DAMDNet(b) and our proposed model(c) on AFLW2000-3D dataset. Face alignment landmarks are shown on RGB images, with a consistent face reconstruction result}
    \label{fig:fig5}
\end{figure}

\begin{figure}
    \centering
    \includegraphics[width=0.9\linewidth]{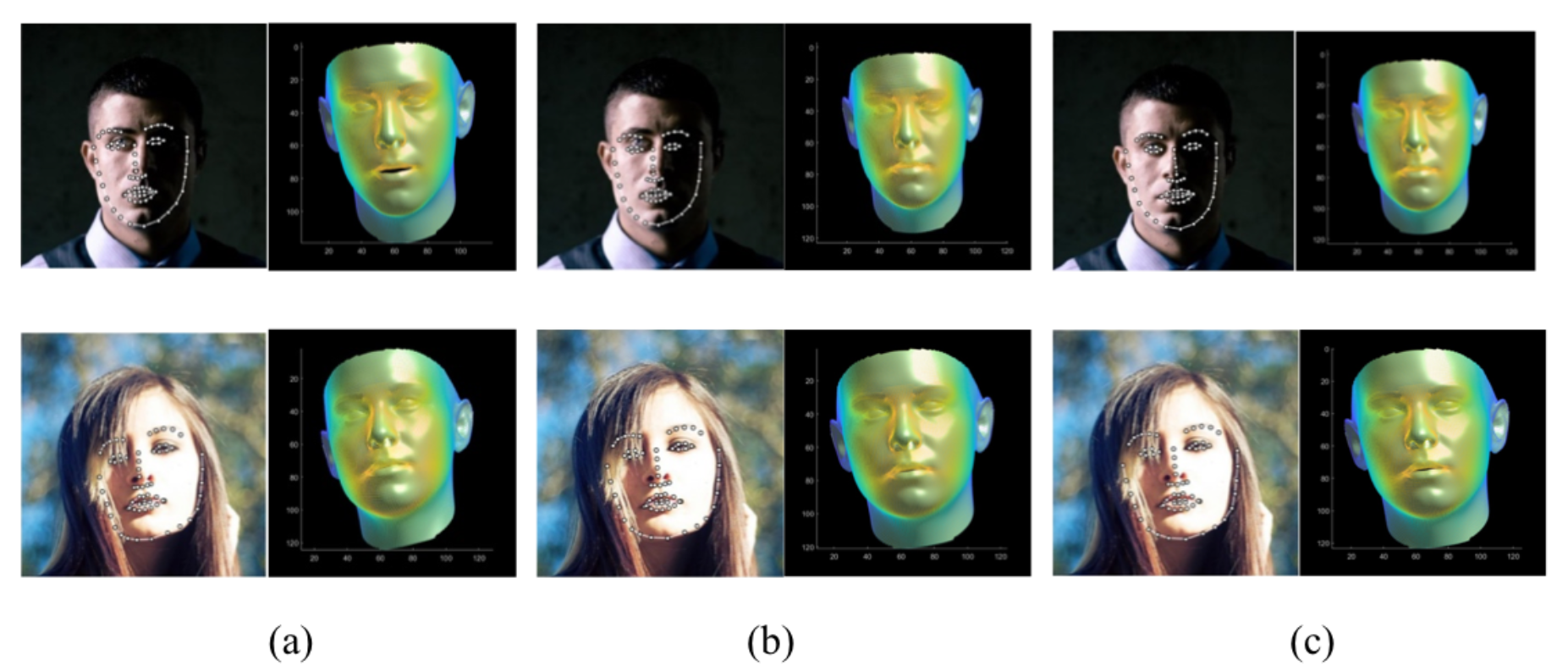}
    \caption{Compared results of 3DDFA, DAMD, our proposed model from left to right}
    \label{fig:fig6}
\end{figure}

For quantitative result, we visualize reconstructed faces on AFLW2000 dataset, as shown in Figure \ref{fig:fig5}.
In face occlusion conditions (e.g. hair-covered face region, incomplete image), our proposed model
can achieve an obviously better result, which captures more accurate face contour, eye corner and nose
tips. The most prominent example can be observed in the figure where a partially present face is also
blocked by sunglasses. Both the 3DDFA model and DAMDNet struggled to generate a reasonable
face shapes, whereas our proposed network successfully detects the face and generates desirable
reconstruction results. Apart from these face details, our model can generate a more reasonable facial
expression and face angles in conditions with large light change Figure \ref{fig:fig6}.

\section{Future Work}
Although our proposed model was able to increase face alignment accuracy on face images with partial occlusion or significant light variation, our model can still be improved to achieve better results. From Figure6 it can be seen that the dark-side face is not accurate enough. First, since the
backbone of our model is inspired by MobileNet \cite{DBLP:journals/corr/HowardZCKWWAA17}, a relatively lightweight network, the ability of our model to capture complex face features is limited. Therefore, increasing the network capacity might be an appropriate step to improve detection accuracy and capture of feature details. Another potential approach is to experiment with loss functions with tunable parameters. Although the weight parameter of the merged loss function can be experimented, the effects did not prove to be very significant. Thus, in future experiments, tunable loss functions may be explored to improve the training of the model.

Our Current method utilizes attention methods to improve robustness against occlusions on a single
image. However, many tasks today require robust 3D face alignment and reconstruction on image
sequences (e.g. videos), whereas our current model does not take the advantage of having multiple
images as input that correspond to the same alignment target. In a sequence of images, the view angle
of the alignment target may change, or the occlusion region may shift from image to image, resulting
in more information provided than any arbitrary image within the sequence. For future work, we
might investigate the possibility of developing a network that focuses on image sequences as inputs,
such as learning a pose-independent 3D face alignment for the target and update the alignment for
every image within the image sequence, to achieve better alignment accuracy and robustness.

\section{Conclusions}
In conclusion, in this project, we investigated the state-of-the-art researches in the field of 3D face
alignment and reconstruction, specifically focusing on 3D face alignment. We first implemented
the 3DDFA network and found that the proposed network can be computationally expensive, timeconsuming and performs relatively unsatisfactory results. Therefore, inspired by recent break
through achieved by attention neural networks, we proposed a network with state-of-the-art attention
mechanisms, namely the Spatial Group-wise Enhancement(SGE) and the Coordinate Attention
Module(CA), to improve the performance of the model when faces in images are partially blocked, or
when lighting in the image drastically changes. We also adopted a joint loss of Weighted Parameter
Distance Cost(WPDC) and Wing Loss to improve convergence and stabilize parameter updates.
Based on the test results on benchmark datasests, our proposed CSANet achieves an improvement on accuracy of 7.44\% on the AFLW dataset and 10.16\% on the AFLW2000-300D dataset. This work
can be applied to generate 3D face scans without human efforts and expensive device.

\bibliographystyle{unsrtnat}


\end{document}